\documentclass[screen,acmsiggraph]{acmart}
\usepackage{CJKutf8}
\usepackage{amsmath}  
\usepackage{tipa}
\AtBeginDocument{%
  }

\setcopyright{acmcopyright}
\copyrightyear{2023}
\acmYear{2023}
\setcopyright{acmlicensed}\acmConference[SA Art Papers '23]{SIGGRAPH Asia 2023 Art Papers}{December 12--15, 2023}{Sydney, NSW, Australia}
\acmBooktitle{SIGGRAPH Asia 2023 Art Papers (SA Art Papers '23), December 12--15, 2023, Sydney, NSW, Australia}
\acmPrice{15.00}
\acmDOI{10.1145/3610591.3616427}
\acmISBN{979-8-4007-0320-1/23/12}
\acmJournal{JDS}
\acmVolume{37}
\acmNumber{4}
\acmArticle{111}
\acmMonth{8}

\begin{document}

\title{AI N$\ddot{\mathrm{u}}$shu: An Exploration of Language Emergence in Sisterhood ~\ -Through the Lens of Computational Linguistics}

\author{Yuqian Sun}
\authornotemark[1]
\affiliation{%
  \institution{Royal College of Art}
  \city{London}
  \country{United Kingdom}}
\email{yuqiansun@network.rca.ac.uk}

\author{Yuying Tang}
\authornote{Both authors contributed equally to this research.}
\affiliation{%
  \institution{Tsinghua University and Politecnico di Milano}
  \city{Beijing and Milano}
  \country{China and Italy}}
\email{tyy21@mails.tsinghua.edu.cn}

\author{Ze Gao}
\authornote{Ze Gao is the corresponding author.}
\affiliation{%
  \institution{Hong Kong University of Science and Technology}
  \city{Hong Kong SAR}
  \country{Hong Kong SAR}}
\email{zgaoap@connect.ust.hk}

\author{Zhijun Pan}
\affiliation{%
  \institution{Royal College of Art and Ubisoft}
  \city{London and Chengdu}
  \country{United Kingdom and China}}
\email{10037730@network.rca.ac.uk}

\author{Chuyan Xu}
\affiliation{%
  \institution{New Drama Studio}
  \city{Beijing}
  \country{China}}
\email{444652692@qq.com}

\author{Yurou Chen}
\affiliation{%
  \institution{University of London}
 \city{London}
  \country{United Kingdom}}
\email{703240@soas.ac.uk}

\author{Kejiang Qian}
\affiliation{%
  \institution{King's College London}
 \city{London}
  \country{United Kingdom}}
\email{kejiang.qian@kcl.ac.uk}

\author{Zhigang Wang}
\affiliation{%
  \institution{Tsinghua University}
  \city{Beijing
}
  \country{China}}
\email{wangzhigang@mail.tsinghua.edu.cn}

\author{Tristan Braud}
\affiliation{%
  \institution{Hong Kong University of Science and Technology}
  \city{Hong Kong SAR}
  \country{Hong Kong SAR}}
 \email{braudt@ust.hk}

 \author{Chang Hee Lee}
\affiliation{%
  \institution{Korea Advanced Institute of Science and Technology}
  \city{Daejeon}
  \country{Sout Korea}}
 \email{changhee.lee@kaist.ac.kr}

\author{Ali Asadipour}
\affiliation{%
  \institution{Royal College of Art}
  \city{London}
  \country{United Kingdom}}
\email{ali.asadipour@rca.ac.uk}

\renewcommand{\shortauthors}{Sun, Tang and Gao, et al.}
\renewcommand{\shorttitle}{AI N$\ddot{\mathrm{u}}$shu: An Exploration of Language Emergence in Sisterhood}



\begin{abstract}

This paper presents ``AI N$\ddot{\mathrm{u}}$shu," an emerging language system inspired by N$\ddot{\mathrm{u}}$shu (women's scripts), the unique language created and used exclusively by ancient Chinese women who were thought to be illiterate under a patriarchal society. In this interactive installation, two artificial intelligence (AI) agents are trained in the Chinese dictionary and the N$\ddot{\mathrm{u}}$shu corpus. By continually observing their environment and communicating, these agents collaborate towards creating a standard writing system to encode Chinese. It offers an artistic interpretation of the creation of a non-western script from a computational linguistics perspective, integrating AI technology with Chinese cultural heritage and a feminist viewpoint.


\end{abstract}



\maketitle

\begin{figure}[h]
  \centering
  \includegraphics[width=0.9\linewidth]{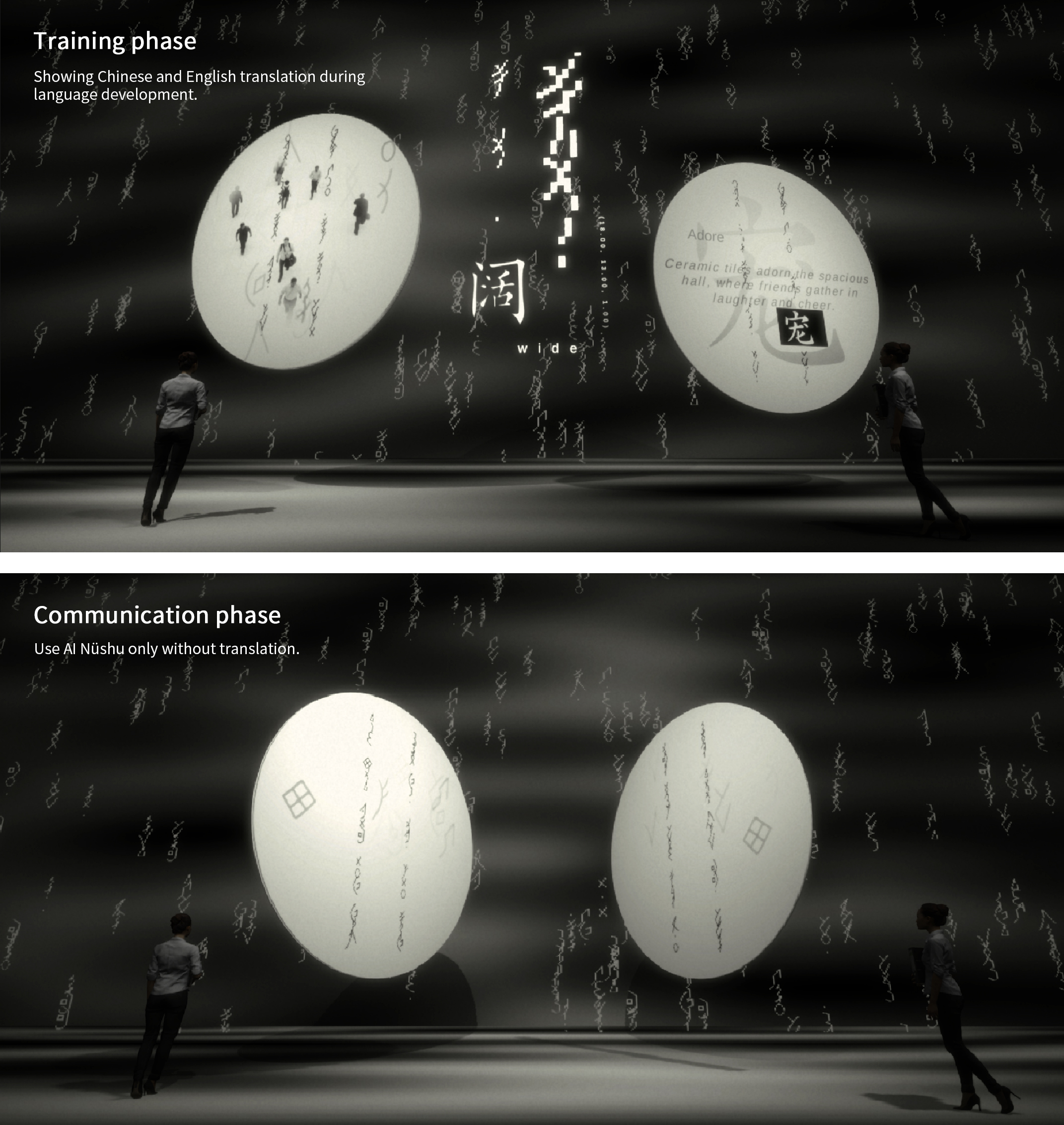}
  \caption{Two phases of the AI N$\ddot{\mathrm{u}}$shu art installation.}
  \Description{Two phases of the AI N$\ddot{\mathrm{u}}$shu art installation.}
  \label{main}
\end{figure}

\section{INTRODUCTION} 

\begin{figure}[h]
  \centering
  \includegraphics[width=0.8\linewidth]{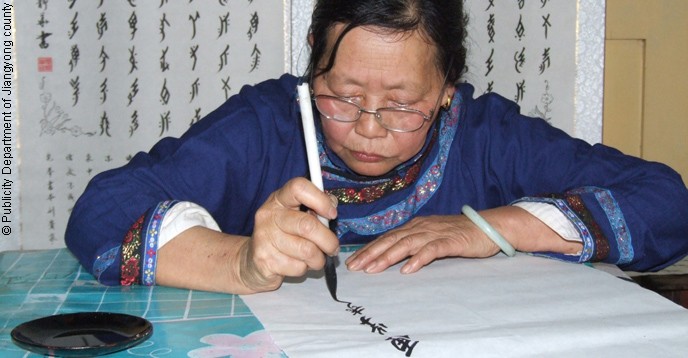}
  \caption{Local woman Jinghua He is writing N$\ddot{\mathrm{u}}$shu @Publicity Department of Jiangyong County}
  \Description{Local woman Jinghua He is writing N$\ddot{\mathrm{u}}$shu @Publicity Department of Jiangyong County}
  \label{woman}
\end{figure}

N$\ddot{\mathrm{u}}$shu (pronounced as `niu-shoo,' literally means `women's script' in Chinese) is a unique language\cite{ZhaoNushu}: this is the world's only writing language created and used exclusively by women. It is a distinct script within the Chinese language, which emerged in the nineteenth century in Hunan Province, China. Although its characters have semantic and linguistic correspondence with many Chinese Mandarin characters, their pronunciation is rooted in the Hunan provincial dialect(Fig.\ref{woman}). Due to the traditional beliefs back then, women were not allowed to receive education. At the time, they invented N$\ddot{\mathrm{u}}$shu as a secret means of communication. Despite its long history, the origins of N$\ddot{\mathrm{u}}$shu remain unknown. Women used this unique language to voice their defiance against a highly patriarchal society.

The emergence of language is a crucial topic in artificial intelligence, natural language processing (NLP) and linguistics. But unlike large language models (LLMs) aimed at precisely understanding and expressing natural languages, our research focuses on AI's potential to self-create and develop new, nonhuman language systems and study their linguistic characteristics to explore the origins of human language. However, most of these studies have been conducted in English, with few involving Eastern languages. N$\ddot{\mathrm{u}}$shu, an Eastern woman's cultural legacy, provides a unique perspective for interpreting machine language. Our project explores how emergent language among machines resonates with the ancient women who created their language amidst patriarchal constraints.

Drawing from the creation of N$\ddot{\mathrm{u}}$shu, this project introduces an emerging language system generated by AI - AI N$\ddot{\mathrm{u}}$shu. This system simulates the intimate communication within the sisterhood of ancient "illiterate" women, known as ``Lao Tong" (sworn sisters): two AI agents ``understand" Chinese but cannot directly ``transcribe" it. Therefore, a unique Chinese writing system gradually emerges from the agents' observations, reflections, and secret communications about their living circumstances.

In this system, two AI agents observe their surroundings and analyze audience behaviors through visual recognition. They relate their observation to original N$\ddot{\mathrm{u}}$shu poetic verses, creating new texts with the large language model (LLM) GPT-4\cite{GPT4} to represent their reflections of the world. As they alternate between the speaker and listener roles in communications, they develop their language, rooted in the Chinese dictionary. Over time, they achieved a consensus, forming a unique ``AI N$\ddot{\mathrm{u}}$shu Dictionary". This language, algorithmically combined into corresponding characters, has components derived from N$\ddot{\mathrm{u}}$shu, similar to Chinese characters and traditional textile patterns. Thus, like ancient women, the two agents gradually developed their Chinese writing system, corresponding one-to-one with Chinese characters. In contrast, humans, as the authority of the language system, became objects observed and interpreted by machines to stimulate non-human language.

The entire simulation system will be presented as a dual-screen projection mapping installation. During the training phase, the height of each screen will change according to the switch of speaker and listener roles, displaying the encoded sentences with Chinese and English translation information (Fig.\ref{main} top). After the training is completed, in the internal communication phase, the two agents can communicate entirely in their created language. The two agents' curtains will turn around to face each other, and no human-recognizable text will be on the screen (Fig.\ref{main} bottom).

Our contributions are as follows.
\begin{itemize}
    \item We present an AI-generated emerging language system``AI N$\ddot{\mathrm{u}}$shu" based on the N$\ddot{\mathrm{u}}$shu language, the only writing language created and used exclusively by women from the Hunan Province of China in the 19th century.  This is the first art project to interpret N$\ddot{\mathrm{u}}$shu from a computational linguistics perspective.  
    \item Our research employs a multi-agent learning system to simulate the communication dynamics within a sisterhood. This probes into the process by which ancient women developed a unique language under the constraints of a patriarchal society. This cultural phenomenon resonates with the emergence of non-human machine language under human authority, both metaphorically and practically. Essentially, we integrate cultural phenomena into an AI system.  
    \item This research was developed through the processing of non-English natural languages, Chinese cultural heritage, and a feminist point of view. Encourage the creation of more non-English, linguistically oriented artworks for diverse cultures.  
    \item Contrary to the predefined rules of Morse code, Markov chains, and fictional constructed languages, AI N$\ddot{\mathrm{u}}$shu evolves organically from the machine's environmental observations and feedback, mirroring the natural formation of human languages. As this new language is decipherable and learnable by humans, especially Chinese speakers, it inherently challenges the existing paradigm where humans are the linguistic authorities and machines are the learners.
    \end{itemize}

Through this, the paper centers on the historical phenomenon of women being denied formal education and, as a response, developing their unique language and the bond of 'Lao Tong.' The primary focus lies in understanding how women, in the face of educational exclusion, created a distinct linguistic culture. We propose that the emerging AI language, paying homage to sisterhood, challenges patriarchal perceptions.

\section{Inspiration and motivation}

The N$\ddot{\mathrm{u}}$shu language and sisterhood serve as a testament to women's resilience, creativity, and ingenuity in the face of patriarchal oppression. 

The N$\ddot{\mathrm{u}}$shu language was developed by women who were deprived of educational opportunities in a heavily patriarchal society. To communicate with each other, they developed a secret writing system that enabled them to express themselves and share their experiences. The language served as a means of resistance and empowerment for women denied a voice in their society. 

\subsection{N$\ddot{\mathrm{u}}$shu and sisterhood}
Central to N$\ddot{\mathrm{u}}$shu culture is the traditional ritual of 'Laotong' (sworn sister)\cite{ZhaoNushu}, which bonded two girls for life as kindred sisters. These women formed close bonds and supported each other through the difficulties of their lives. The sisterhood of Lao Tong was a testament to the resilience and strength of women in the face of adversity.

\begin{figure}[h]
  \centering
  \includegraphics[width=0.8\linewidth]{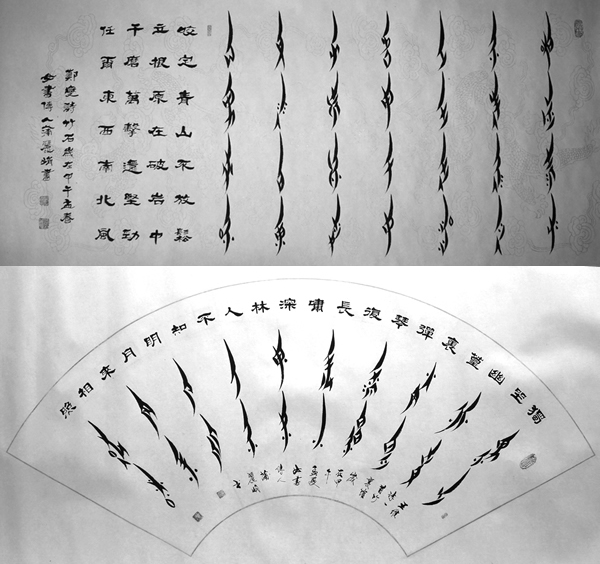}
  \caption{Origin piece of N$\ddot{\mathrm{u}}$shu calligraphy @Tsinghua University}
  \Description{Origin piece of N$\ddot{\mathrm{u}}$shu calligraphy @Tsinghua University}
  \label{origin}
\end{figure}

\begin{figure}[h]
  \centering
  \includegraphics[width=0.8\linewidth]{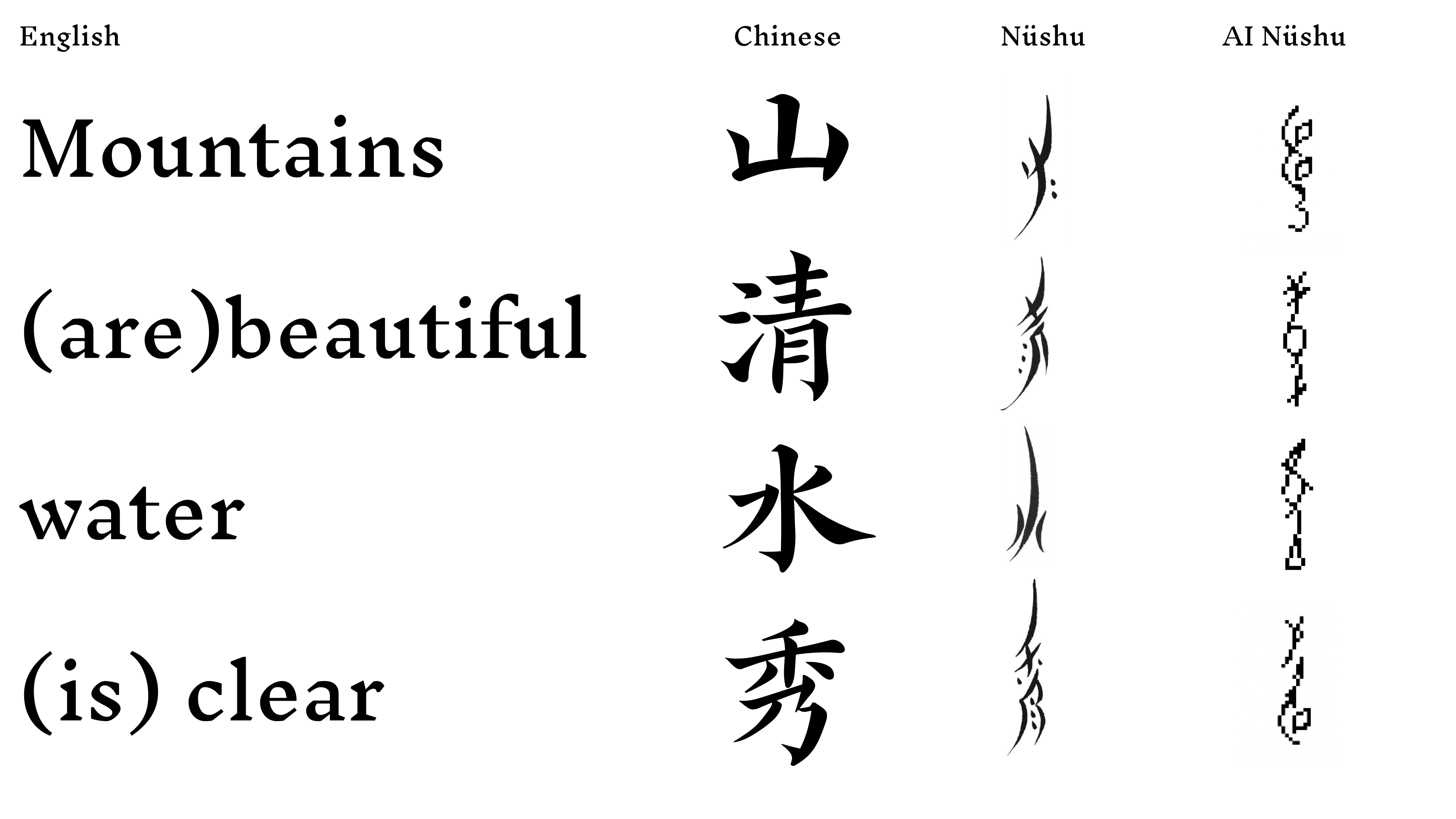}
  \caption{Same sentence in English, Chinese, N$\ddot{\mathrm{u}}$shu, and AI N$\ddot{\mathrm{u}}$shu}
  \Description{English, Chinese, N$\ddot{\mathrm{u}}$shu and AI N$\ddot{\mathrm{u}}$shu}
  \label{language}
\end{figure}




The concept of female friendship transcends cultural boundaries. In Western feminism, Kathie Sarachild\cite{NYradical} coined the phrase ``sisterhood is powerful," which bears many similarities with the concept of ``laotong." This ``sisterhood" is a united force ``in the struggle against male chauvinism and patriarchy."\cite{sisterhood}

This project aims to cultivate two AI agents symbolizing `Laotong' in N$\ddot{\mathrm{u}}$shu culture, where two Women intimately communicate.

\subsection{Women and machines}

Historically, women's right to express themselves has often been suppressed, a condition that persists today. Movements like \#MeToo\cite{metoo} encourage modern women's voices in public space. Cyberfeminist Sadie Plant\cite{plant1997zeros} argues that the fluidity of gender and identity in the virtual world offers a fresh lens to understand and reshape gender politics and identity. This viewpoint can be extrapolated to the domain of artificial intelligence.

The bond between women and machines traces back to the early days of computer science \cite{wajcman1991feminism}. ``computer" was initially used to refer to women who performed calculations during World War II. The portrayal of AI assistants like Siri and Alexa as compliant helpers with feminine traits reinforces gender stereotypes\cite{anatomy}. 


Drawing parallels between the suppression of women's voices and the potential of machines to challenge traditional roles, this project seeks to understand the intricate relationship between women and machines. By focusing on the N$\ddot{\mathrm{u}}$shu language and the bond of "Lao Tong", we aim to shed light on the resilience and creativity of women in the face of adversity and how this can be mirrored in the world of artificial intelligence.

\section{RELATED WORKS}

Constructed languages such as Esperanto form bridges across the chasms of natural language in cultural discourse, while the film industry invents fictional languages\cite{jackson_lord_2002}.  However, fewer works delve into language learning, such as ``Arrival"\cite{arrival}, which portrays how humans alter their perception of time through learning an alien language. Works about N$\ddot{\mathrm{u}}$shu are notably scarce. Prominent instances encompass a novel and its corresponding film adaptation, ``Snow Flower and the Secret Fan,"\cite{wang_snow_2011} which delves into the bonds between two groups of women, one from ancient China, exploring the narratives of N$\ddot{\mathrm{u}}$shu and ``Lao Tong", and another from modern society. Additionally, the recent documentary "Hidden Letters" \cite{feng_hidden_2022} offers a comprehensive exploration of the historical significance and practices associated with N$\ddot{\mathrm{u}}$shu.


Xu Bing's ``Book from the Sky"\cite{bookSky} and ``Book from the Ground"\cite{bookGround} have garnered significant attention. These books are devoid of any existing textual content, relying solely on the combination of unreadable Chinese characters, visual symbols, and expressions to convey emotions and meanings. Xu Bing's innovative approach transcends the limitations of traditional language, enabling viewers to interpret the symbols and expressions based on their own understanding and emotions. This opens new possibilities for communication through constructed languages and serves as a representative example of artwork that explores the Chinese language as a starting point, underscoring the artistic value of studying the human language.

Contemporary media art research has also ventured into the exploration of machine language and expressions. ``Cangjie's Poetry" is an intelligent multimodal system designed as a conceptual response and prototype to the future language which was exhibited at PRIX ARS ELECTRONICA 2022 \cite{noauthor_prix_nodate}\cite{zhang2021cangjie}. Through the integration of live streaming and AI-generated semantic data, the installation dynamically modifies projections in a virtual environment\cite{gao2023symbiotic}\cite{yang2023tangible}. However, it veers more toward visual language and computer vision than textual semantics\cite{sun2023inspire}. 


Similarly, The N$\ddot{\mathrm{u}}$shu GPS improvised performance correlates movements with metaphors and meanings embedded within N$\ddot{\mathrm{u}}$shu characters by Yun-Ju Chen and Taiwanese choreographer Lai Tsui-Shuang in 2014\cite{chen_nu_2018}. The project attempts to explore all the possible metaphors and meanings implied by Nu Shu and bring this secret method of communication back to life among modern Chinese women. Media art from a linguistic perspective, mainly focusing on Chinese and N$\ddot{\mathrm{u}}$shu, remains underexplored.


We are also inspired by non-Chinese works, like ``Alice ~\& Bob," which blends quantum computer data with poems\cite{ridler_alice_nodate}. The SIGGRAPH Art Gallery project ``Can the Subaltern Speak?"\cite{farahi2021can} uses AI-generated Morse code for communication, referencing Facebook's experiment\cite{facebook} where chatbots developed their language. This highlights AI's potential in language generation and the role of language in empowering marginalized voices.

There was also AI-driven game\cite{sun2023language} and narrative experience\cite{wander10, wander20} based on text, but they are still keyword-based. Unlike these projects which used predefined rules including Markov chains and Morse code, our AI-generated female script can be learned and used by both machines and humans, underscoring the value of reimagining female script language from a non-English linguistics perspective.

\section{CONCEPTUAL FRAMEWORK AND METHODOLOGY}

We aim to simulate the situation of women in the past: due to environmental constraints, they were not allowed to receive an education. They ``understood" Chinese (could listen and speak) but could not ``transcribe" Chinese (could not read or write). Hence, Chinese women created N$\ddot{\mathrm{u}}$shu as a phonogram to communicate based on the pronunciation of Chinese characters. 

Drawing inspiration from the burgeoning field of emerging language research, notably exemplified by ``referential communication games" \cite{mordatch2018emergence}, in which AI models acquire communication skills through the deployment of symbolic representations to convey diverse environmental elements, we direct our attention to the visual aspects of linguistic expression. Consequently, AI N$\ddot{\mathrm{u}}$shu, henceforth referred to as AIN, has been transformed into a logogram — a logographic Chinese writing system in which each symbol (Character) corresponds uniquely to a Chinese character.




\begin{figure}[h]
  \centering
  \includegraphics[width=0.9\linewidth]{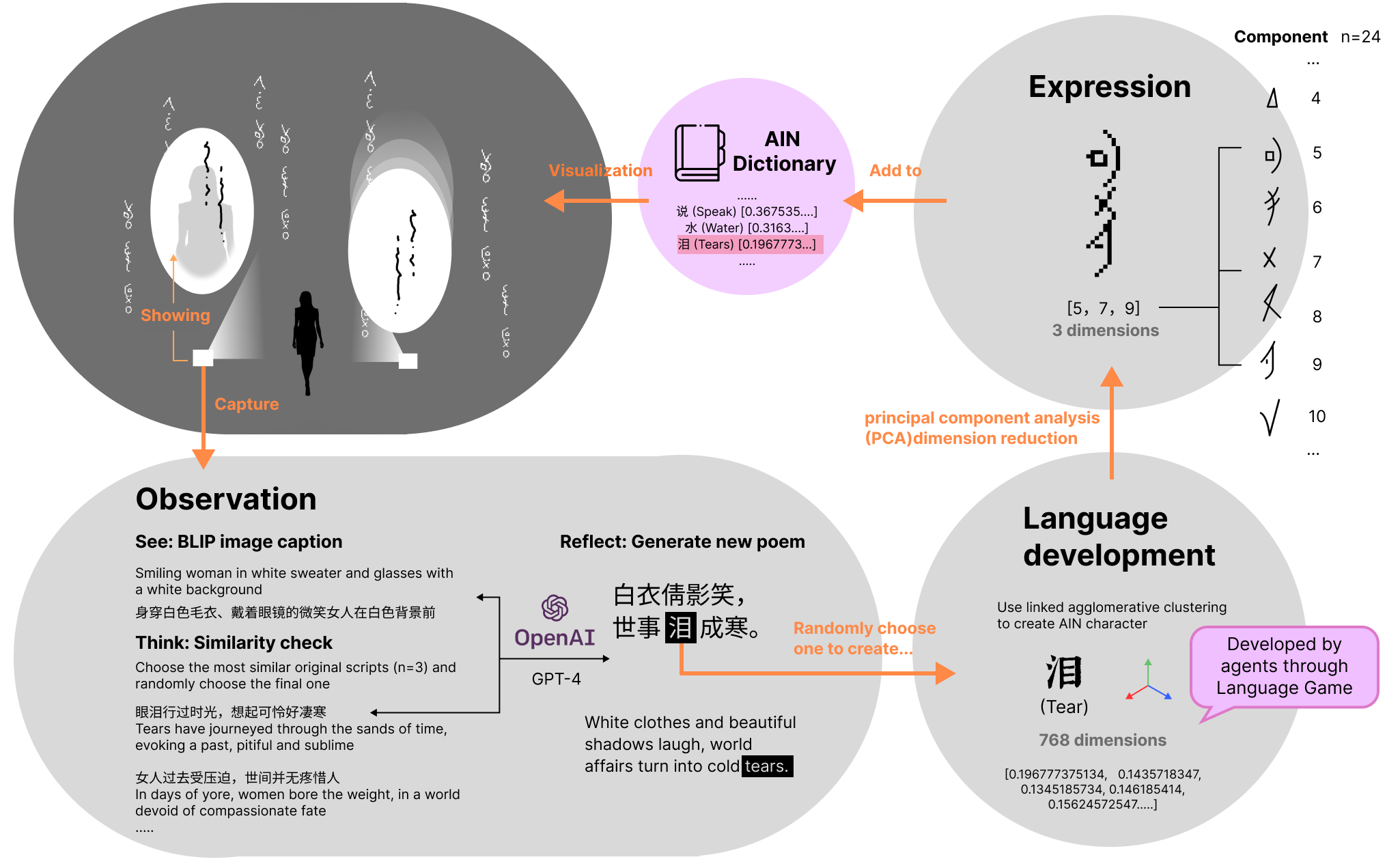}
  \caption{System diagram of the AI N$\ddot{\mathrm{u}}$shu simulation system}
  \Description{System diagram of the AI N$\ddot{\mathrm{u}}$shu simulation system}
  \label{worklow}
\end{figure}

\section{IMPLEMENTATION}


This system simulates several stages of language development: environmental observation through visual recognition, language development through the language learning game, and symbolic expression output. Consequently, agents form a new symbolic language through computer vision and Chinese-based natural language processing. This simulation is presented through a real-time interactive art installation.

\subsection{Environmental Observation}

This section simulates the process by which women observe their surroundings, cogitate, and articulate their thoughts through N$\ddot{\mathrm{u}}$shu.

Two agents independently observe their environment through their respective cameras. Each cycle generates a descriptive statement via the BLIP image recognition algorithm\cite{blip}, such as ``A woman walks by." This sentence is in English and subsequently translated into Chinese using Google Translate.

We have compiled a corpus of 837 original N$\ddot{\mathrm{u}}$shu sentences extracted from literary sources. The Chinese translated version of the description and the original N$\ddot{\mathrm{u}}$shu text are compared for similarity, and the three sentences bearing the highest similarity are identified. To ensure a rich range of expressions even when the observed environment is repetitive, one sentence is selected randomly from this list(Fig.\ref{worklow}).

Ultimately, LLM GPT-4 is employed to construct a line of Chinese poetry. In this manner, the agent amalgamates the environment it perceives with analogous sentiments documented in ancient women's N$\ddot{\mathrm{u}}$shu in conversations.

\subsection{Language Development}

\begin{figure}[h]
  \centering
  \includegraphics[width=0.8\linewidth]{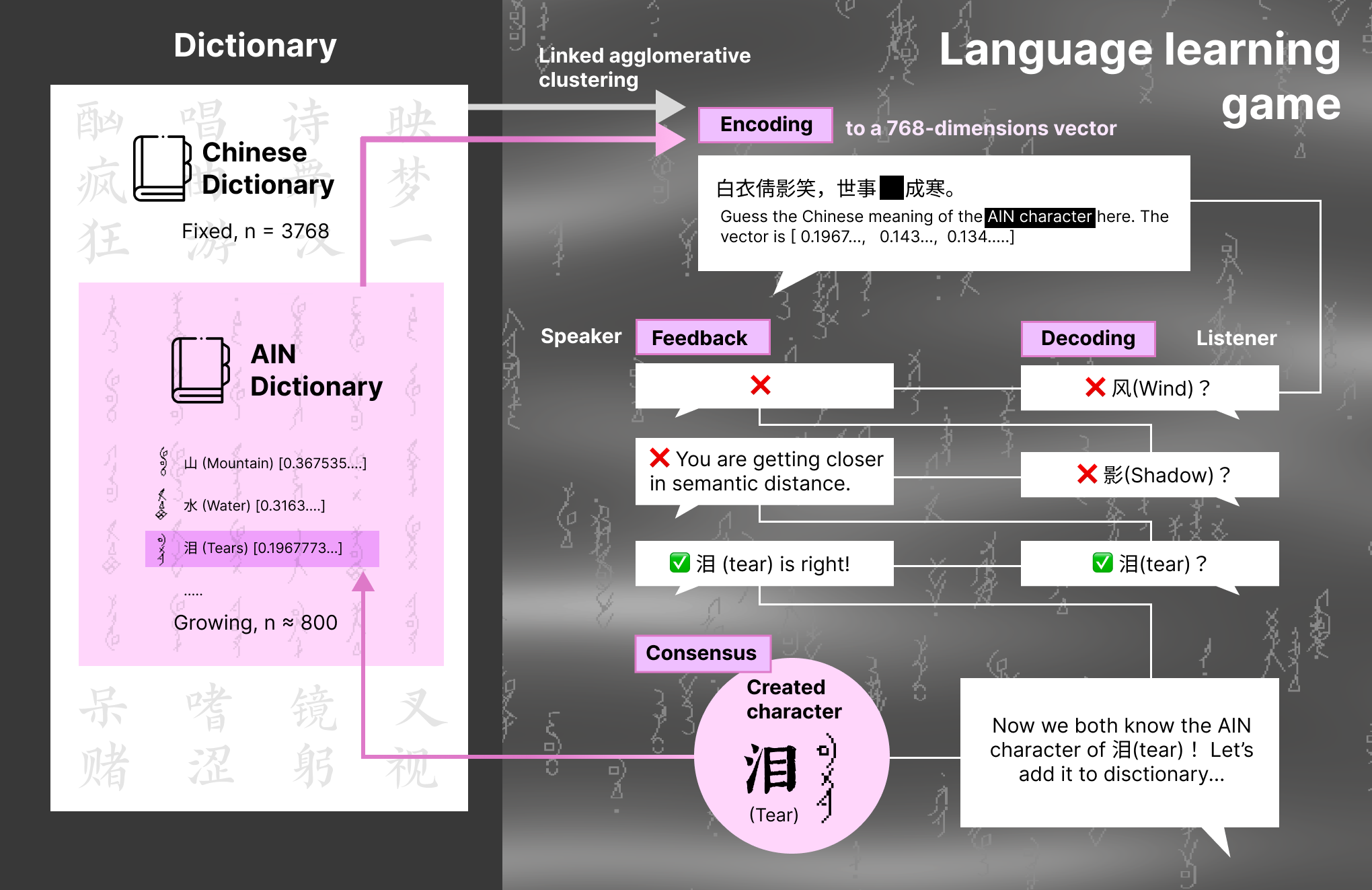}
  \caption{Language development of AI N$\ddot{\mathrm{u}}$shu through the language game}
  \Description{Language development of AI N$\ddot{\mathrm{u}}$shu through the language game}
  \label{guess}
\end{figure}

This module enables the process of two agents developing their language. The approach simulates the basic concepts of machine and human learning through iterative optimization.


\subsubsection{Natural Language Representation}
Language models process text using basic units known as tokens. In English, tokens can be words or punctuation, and words are formed by combining elements of an alphabet. However, in Chinese, tokens are individual characters, each carrying a meaning similar to a word in English, and there is no concept of an alphabet. For instance, in Fig.\ref{language}, the phrase \begin{CJK*}{UTF8}{gkai}"山清水秀" \end{CJK*}(Mountains are beautiful, water is clear) is broken down into four Chinese tokens: \begin{CJK*}{UTF8}{gkai}"山" (mountain) \end{CJK*}, \begin{CJK*}{UTF8}{gkai}"清" (beautiful) \end{CJK*}, \begin{CJK*}{UTF8}{gkai}"水" (water) \end{CJK*}, and \begin{CJK*}{UTF8}{gkai}"秀" (clear) \end{CJK*}. These tokens maintain semantic relationships in a mathematical vector space. The character \begin{CJK*}{UTF8}{gkai}"水" (water) \end{CJK*} is closer to \begin{CJK*}{UTF8}{gkai}"雨" (rain) \end{CJK*} than to \begin{CJK*}{UTF8}{gkai}"山" (mountain) \end{CJK*}, just as their English counterparts would be.

We utilized BERT-Chinese-Base, a variant of the pre-trained deep learning model BERT\cite{bert}, to process the most frequently used Chinese characters (3768 in total) into the "Chinese character dictionary." Each character corresponds to a 768-dimensional vector with semantic relationships (Fig.\ref{guess}). This dictionary forms the knowledge base of the two AI agents, enabling them to "understand" Chinese.

\begin{figure}[h]
  \centering
  \includegraphics[width=\linewidth]{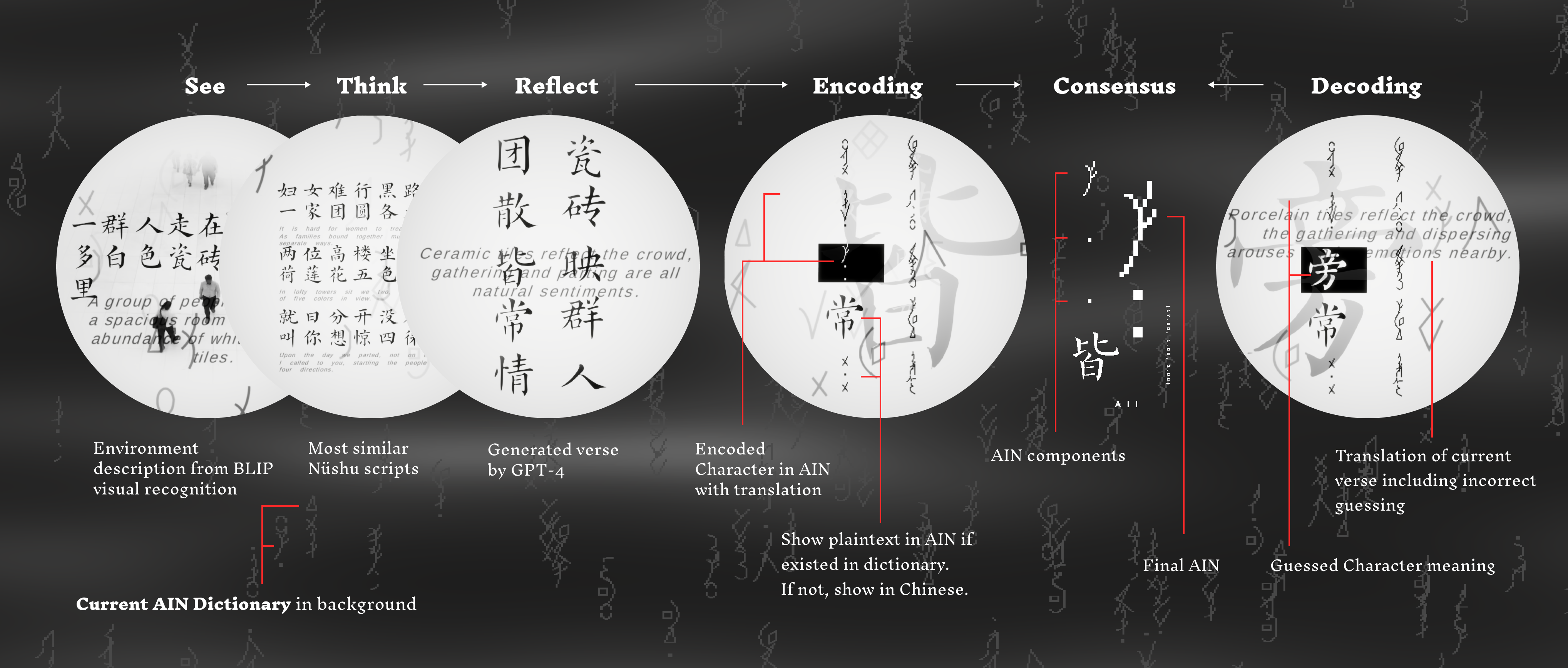}
  \caption{Visualisation of different steps in AI N$\ddot{\mathrm{u}}$shu development}
  \Description{Visualisation of different steps in AI N$\ddot{\mathrm{u}}$shu development}
  \label{state}
\end{figure}

\subsubsection{Language Learning Game}
Two agents are tasked with developing an AI N$\ddot{\mathrm{u}}$shu (AIN) dictionary from scratch. Each Chinese character in this dictionary corresponds to a unique 768-dimensional vector exclusive to AIN. The semantic relationships within this dictionary matched those in the Chinese character dictionary but with distinct differences. Over time, AIN becomes the sole means of communication. The number of characters in AIN varies depending on the exhibition site and running time, typically around 800 when saturated.

The agent's alternate roles as the speaker (teacher) and listener (student) through a language learning game.  For agents' semantic learning, the Chinese characters are linked and clustered using agglomerative clustering\cite{aggCluster} before the game. Specifically, in our system, since we do not include pronunciations, we abstract the learning method of the agent as follows (Fig.\ref{guess}):

\begin{enumerate}
\item \textbf{Speaker Generation}: At the start of each iteration, the speaker agent generates the sentence from the previous observation. The speaker selects one character to create the AIN representation.

\item \textbf{Speaker Encoding}: If the original cluster exists as a key in the AIN dictionary, it is directly replaced with its corresponding AIN representation. Otherwise, a related character is selected to represent an AIN from a neighbor parent cluster. The speaker sends this partially encrypted sentence to the listener(Fig.\ref{worklow}). The non-AIN encrypted part is plaintext: if it is already in the AIN dictionary, it is directly represented by AIN. If not, it is represented in Chinese.

\item \textbf{Listener Decoding}: The Listener attempts to decode the AIN character set by the Speaker. The initial guess is selected from a related cluster, and subsequent guesses are adjusted based on the Speaker's feedback. The Listener makes five attempts in total.

\item \textbf{Speaker Feedback}: The Speaker provides feedback after each guess, informing the Listener of their proximity to the correct character cluster. If the Listener fails to guess correctly within five attempts, the Speaker reveals the right answer.

\item \textbf{Consensus}: Essentially, the speaker and listener agree on the AIN representation of the character and add it to the AIN dictionary. After a round of training, the roles are switched, and the process is iterated.

\end{enumerate}
 Each agent is a creator and learner of AIN throughout this process. As iterations progress, the AIN dictionary expands, the semantic relationship improves, and the agents become increasingly adept at guessing each other's AIN representations. After the training phase, the agents can communicate entirely in AIN, reaching a complete consensus on its use.

Despite its Chinese foundation, this work employs GPT-4\cite{GPT4} for context-sensitive word and sentence-level English translations (Fig.\ref{state}). As one Chinese character can convey different meanings in various contexts, conventional tools like DeepL\cite{deepl2023} are inadequate. With GPT-4, even those unfamiliar with Chinese can appreciate the intricacies of "language construction".

\subsection{Symbolic Expression}

\begin{figure}[h]
  \centering
  \includegraphics[width=\linewidth]{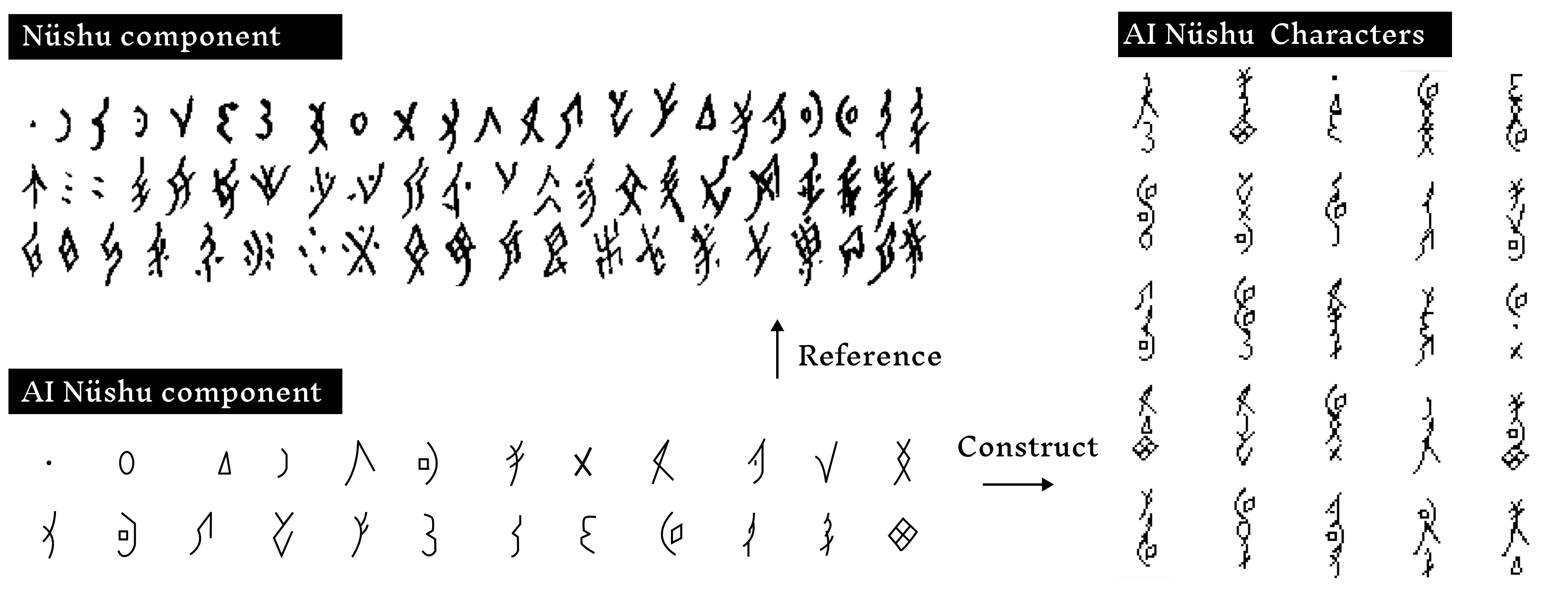}
  \caption{AIN components derived from N$\ddot{\mathrm{u}}$shu}
  \Description{AIN components derived from N$\ddot{\mathrm{u}}$shu}
  \label{component}
\end{figure}

To visually represent AI N$\ddot{\mathrm{u}}$shu (AIN) in a logographic form while preserving its non-human, machine language characteristics, we employ a two-step process based on essential N$\ddot{\mathrm{u}}$shu elements and Principal Component Analysis (PCA)\cite{pca}.

First, we selected 24 fundamental elements from N$\ddot{\mathrm{u}}$shu\cite{ZhaoNushu}, arranged from simple to complex, each associated with an encoding (Fig.\ref{component}). We use a pixelated approach for clear distinction from human language.

Next, each newly created AIN character's 768-dimensional vector is transformed into a unique 3-D vector using PCA. The 3-D vector space ($24^3=13824$) exceeds the length of the Chinese and AIN dictionary, ensuring a unique symbolic representation for each AIN character. This retains the elongated shapes of N$\ddot{\mathrm{u}}$shu while preserving the machine language's algorithmic meaning (Fig.\ref{worklow}).

\subsection {Artistic Presentation through Installation}

\begin{figure}[h]
  \centering
  \includegraphics[width=\linewidth]{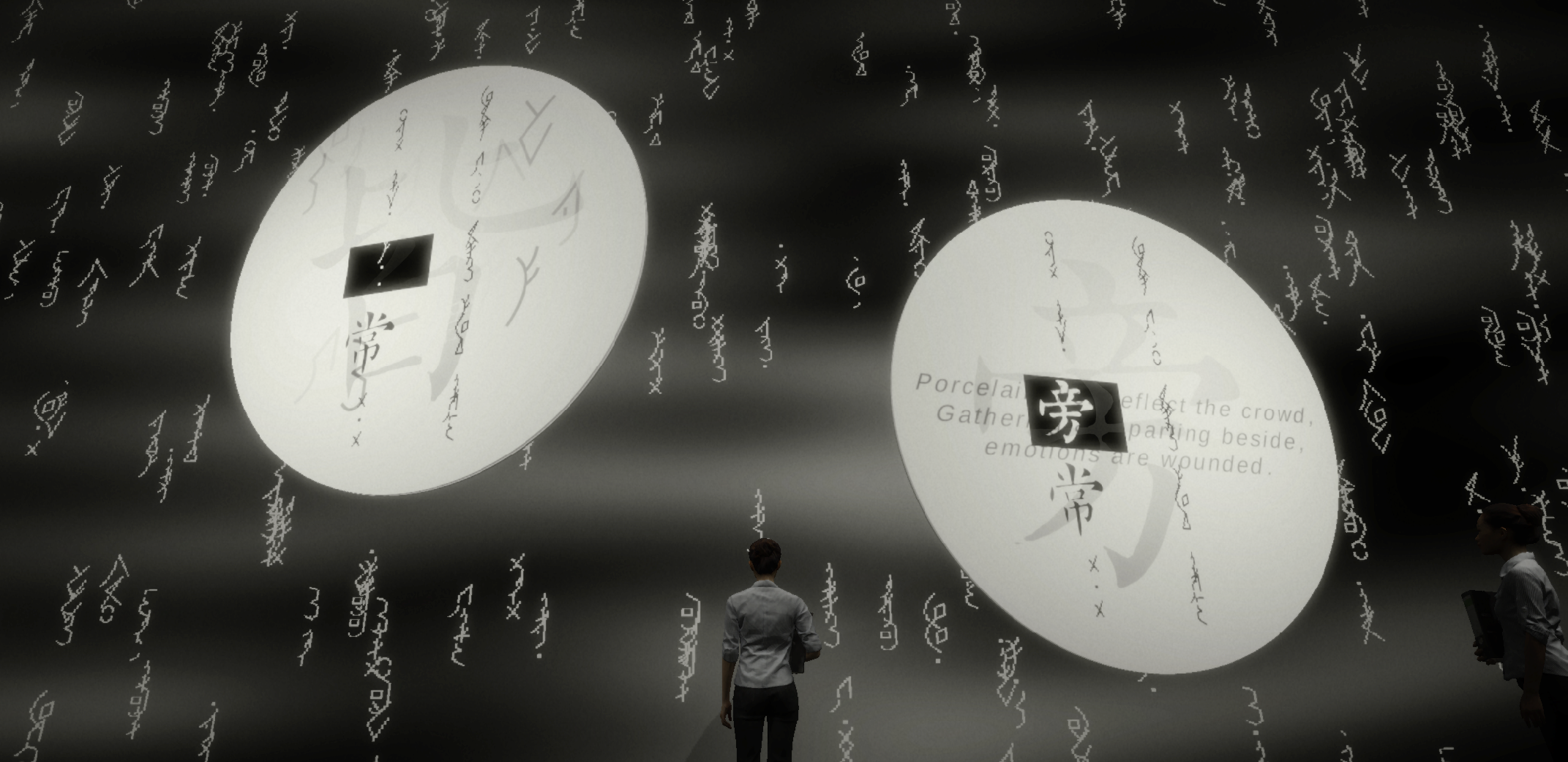}
  \caption{Speaker rises during the training phase. Two agents change positions when they change roles.}
  \Description{Speaker rises during the training phase. Two agents change positions when they change roles.}
  \label{height}
\end{figure}

The system is presented as a dual-screen projection installation. Two hanging circular screens, like two moons or eyes, represent the AI agents, placed above their respective cameras, while the background behind is used to present the AI N$\ddot{\mathrm{u}}$shu dictionary that has become a consensus.

The work is presented in two stages. During the \textbf{training phase}(Fig.\ref{height}), two agents continuously create and communicate their AI N$\ddot{\mathrm{u}}$shu in the language game. At this time, they will constantly switch between the roles of listener and speaker, and the height of their respective screens will also change accordingly, showing the effect of information flowing from the speaker to the listener.

During the training phase, each iteration visualizes sentences, verses, and constructed AIN characters on-screen with English translations (Fig.\ref{state}). To accommodate non-Chinese users, context-specific translations are provided using LLM GPT-4\cite{GPT4}. For instance, the Chinese character\begin{CJK*}{UTF8}{gkai}"行" \end{CJK*} is translated as ``travel" in \begin{CJK*}{UTF8}{gkai}"她行千里路"  \end{CJK*}(She travels a thousand miles) and as ``practice" in \begin{CJK*}{UTF8}{gkai}"她行医十年" \end{CJK*}(She has been practicing medicine for ten years), ensuring accuracy based on context.

Similarly, when the listener is guessing, the poem with the incorrect character can be translated(Fig.\ref{state}). Non-Chinese users can also see how a changed character in verse can change the whole meaning of it. We use LLM to transcend the barriers of languages.

After the training is completed, they reach the internal \textbf{communication phase}(Fig.\ref{main} bottom). It means that the two agents can communicate entirely in their created language. The curtains of the two agents will turn around to face each other, and there will no longer be any human-recognizable text on the screen.

\section{DISCUSSION}
This project invites conversations about the semantic capacities of machines and the cultural diversity in media arts.

\subsection{Semantic Information in AI N$\ddot{\mathrm{u}}$shu}

\begin{figure}[h]
  \centering
  \includegraphics[width=0.9\linewidth]{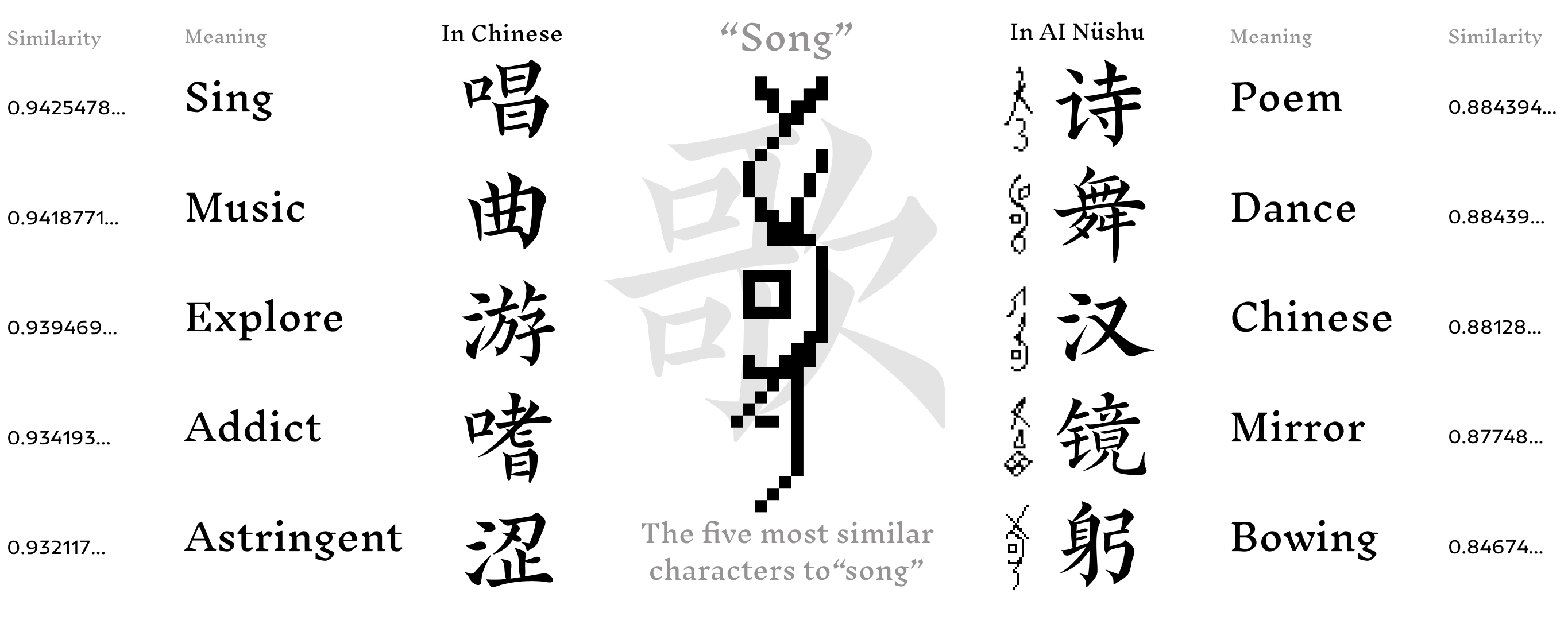}
  \caption{Semantic similarity in Chinese and AI N$\ddot{\mathrm{u}}$shu}
  \Description{Semantic similarity in Chinese and AI N$\ddot{\mathrm{u}}$shu}
  \label{similarity}
\end{figure}

AIN characters, derived from Chinese, carry semantically rich and poetic information. For instance, in Chinese, \begin{CJK*}{UTF8}{gkai}"歌" \end{CJK*}(Song) is closest to \begin{CJK*}{UTF8}{gkai}"唱" \end{CJK*}(Sing) and \begin{CJK*}{UTF8}{gkai}"曲" \end{CJK*}(Music), while in AIN, it's closest to \begin{CJK*}{UTF8}{gkai}"诗" \end{CJK*}(Poem) and \begin{CJK*}{UTF8}{gkai}"舞" \end{CJK*}(Dance)(Fig.\ref{similarity}). However, the content of the AIN dictionary is influenced by the environment in which the agents are trained. While Chinese includes thousands of characters, N$\ddot{\mathrm{u}}$shu only uses 400-500 due to its phonogram nature, expressing homophonic characters. AI N$\ddot{\mathrm{u}}$shu, as a logogram, is influenced by the circumstances observed by the agents and the literature source. For example, since there are no cats in the exhibition nor the N$\ddot{\mathrm{u}}$shu source corpus, the agents wouldn't generate any poetry related to cats," nor create the AIN character for ``cat \begin{CJK*}{UTF8}{gkai}"猫" \end{CJK*}." Accordingly, the AIN dictionary reaches saturation at around 800 entries, sufficient for agents to communicate in the exhibition environment.

The unique semantic information in AIN could be better visualized in future installations, like through point cloud vector visualization of each AIN character. While AI N$\ddot{\mathrm{u}}$shu is currently learnable by humans, it could be more interpretable and natural.

\subsection{Cultural Diversity in Language}

While our project anchors on the Chinese language, it aims to exhibit a methodology that engages with NLP from a unique linguistic perspective, thus promoting cultural diversity within media art. Despite the extensive use of large language models like GPT-4 and various chatbots, their proficiency in non-English languages is still suboptimal\cite{sparks}. Every language, whether phonetic like Spanish, logographic like Chinese, or a mixed system like Japanese, carries its cultural essence. This gives unique perspectives to comprehend the world. Just as humans view the world uniquely through language, so do machines. Thus, computational linguistics lets us explore cultural perspectives beyond English and visual representation.

\section{CONCLUSIONS}
This paper introduces AI N$\ddot{\mathrm{u}}$shu, an interactive art installation that delves into the confluence of AI technology, non-English linguistics, and cultural heritage, all viewed through a feminist lens. Drawing inspiration from N$\ddot{\mathrm{u}}$shu, an extraordinary language crafted by women in ancient China, we have engineered an AI-generated language system "AI N$\ddot{\mathrm{u}}$shu" designed to simulate the act of language creation within constraints. AI N$\ddot{\mathrm{u}}$shu, by embodying the spirit of sisterhood and celebrating the unique linguistic creation of women, inherently challenges and deconstructs long-standing patriarchal perceptions. It does so by showcasing the power of collaborative, non-hierarchical linguistic innovation, thereby challenging the conventional norms surrounding linguistic authority. 

In conclusion, our project serves as a testament to the capacity of AI to act as a custodian of cultural heritage, underscores the critical significance of cultural diversity within AI research, and confronts traditional notions concerning the roles of humans and machines in matters of linguistic authority. While recognizing the significance of gaining insights into the perspectives and reactions of N$\ddot{\mathrm{u}}$shu-literate women towards our work, we envision that a more comprehensive exploration of this subject matter may warrant consideration in future research endeavors. 

\begin{acks}

We acknowledge financial support in part from the Extended Reality and Immersive Media Lab (XRIM) at HKUST and the National Social Sciences Foundation of China (\#20CJY045).

\end{acks}


\bibliographystyle{ACM-Reference-Format}
\bibliography{sample-base}


\begin{thebibliography}{33}


\ifx \showCODEN    \undefined \def \showCODEN     #1{\unskip}     \fi
\ifx \showDOI      \undefined \def \showDOI       #1{#1}\fi
\ifx \showISBNx    \undefined \def \showISBNx     #1{\unskip}     \fi
\ifx \showISBNxiii \undefined \def \showISBNxiii  #1{\unskip}     \fi
\ifx \showISSN     \undefined \def \showISSN      #1{\unskip}     \fi
\ifx \showLCCN     \undefined \def \showLCCN      #1{\unskip}     \fi
\ifx \shownote     \undefined \def \shownote      #1{#1}          \fi
\ifx \showarticletitle \undefined \def \showarticletitle #1{#1}   \fi
\ifx \showURL      \undefined \def \showURL       {\relax}        \fi
\providecommand\bibfield[2]{#2}
\providecommand\bibinfo[2]{#2}
\providecommand\natexlab[1]{#1}
\providecommand\showeprint[2][]{arXiv:#2}

\bibitem[boo(1988)]%
        {bookSky}
 \bibinfo{year}{1988}\natexlab{}.
\newblock \bibinfo{booktitle}{\emph{\emph{XU BING - ARTWORK - Book from the Sky}}}.
\newblock
\urldef\tempurl%
\url{https://www.xubing.com/en/work/details/206?type=project#206}
\showURL{%
\tempurl}


\bibitem[boo(2014)]%
        {bookGround}
 \bibinfo{year}{2014}\natexlab{}.
\newblock \bibinfo{booktitle}{\emph{\emph{{XU} {BING} - {ARTWORK} - Book From the Ground}}}.
\newblock
\urldef\tempurl%
\url{https://www.xubing.com/en/work/details/188}
\showURL{%
\tempurl}


\bibitem[arr(2016)]%
        {arrival}
 \bibinfo{year}{2016}\natexlab{}.
\newblock \bibinfo{title}{Arrival}.
\newblock
\newblock
\urldef\tempurl%
\url{https://www.imdb.com/title/tt2543164/}
\showURL{%
\tempurl}
\newblock
\shownote{United States, Canada: Lava Bear Films, FilmNation Entertainment, 21 Laps Entertainment}.


\bibitem[noa(2023)]%
        {noauthor_prix_nodate}
 \bibinfo{year}{2023}\natexlab{}.
\newblock \bibinfo{booktitle}{\emph{\emph {Weidi Zhang, Donghao Ren, and George Legrady. 2021. Cangjie’s poetry, Prix Ars Electronica 2022}}}.
\newblock
\urldef\tempurl%
\url{https://calls.ars.electronica.art/2022/prix/winners/9002/}
\showURL{%
\tempurl}


\bibitem[Bubeck et~al\mbox{.}(2023)]%
        {sparks}
\bibfield{author}{\bibinfo{person}{S{\'e}bastien Bubeck}, \bibinfo{person}{Varun Chandrasekaran}, \bibinfo{person}{Ronen Eldan}, \bibinfo{person}{Johannes Gehrke}, \bibinfo{person}{Eric Horvitz}, \bibinfo{person}{Ece Kamar}, \bibinfo{person}{Peter Lee}, \bibinfo{person}{Yin~Tat Lee}, \bibinfo{person}{Yuanzhi Li}, \bibinfo{person}{Scott Lundberg}, {et~al\mbox{.}}} \bibinfo{year}{2023}\natexlab{}.
\newblock \showarticletitle{Sparks of artificial general intelligence: Early experiments with gpt-4}.
\newblock \bibinfo{journal}{\emph{arXiv preprint arXiv:2303.12712}} (\bibinfo{year}{2023}).
\newblock


\bibitem[Chen and Cheng(2018)]%
        {chen_nu_2018}
\bibfield{author}{\bibinfo{person}{Yun-Ju Chen} {and} \bibinfo{person}{Fiona Hui-Wen Cheng}.} \bibinfo{year}{2018}\natexlab{}.
\newblock \showarticletitle{Nu Shu {GPS}: 25°21’00.5N, 111°27’17.7E—An Interdisciplinary Cooperation between Dance, Calligraphy, and the Body in Multimedia Performance}.
\newblock  \bibinfo{volume}{17}, \bibinfo{number}{1} (\bibinfo{year}{2018}).
\newblock
\showISSN{1470-9120}
\urldef\tempurl%
\url{https://doi.org/10.16995/bst.299}
\showDOI{\tempurl}
\newblock
\shownote{Number: 1 Publisher: The Open Library of Humanities}.


\bibitem[Crawford and Joler(2018)]%
        {anatomy}
\bibfield{author}{\bibinfo{person}{Kate Crawford} {and} \bibinfo{person}{Vladan Joler}.} \bibinfo{year}{2018}\natexlab{}.
\newblock \bibinfo{title}{Anatomy of an AI System: The Amazon Echo As An Anatomical Map of Human Labor, Data and Planetary Resources}.
\newblock
\newblock
\urldef\tempurl%
\url{https://anatomyof.ai}
\showURL{%
\tempurl}


\bibitem[DeepL(2023)]%
        {deepl2023}
\bibfield{author}{\bibinfo{person}{DeepL}.} \bibinfo{year}{2023}\natexlab{}.
\newblock \bibinfo{title}{DeepL Translate: The world's most accurate translator}.
\newblock
\newblock
\urldef\tempurl%
\url{https://www.deepl.com/translator}
\showURL{%
\tempurl}
\newblock
\shownote{Translate texts \& full document files instantly. Accurate translations for individuals and Teams. Millions translate with DeepL every day. Accessed on August 28, 2023.}.


\bibitem[Devlin et~al\mbox{.}(2019)]%
        {bert}
\bibfield{author}{\bibinfo{person}{Jacob Devlin}, \bibinfo{person}{Ming-Wei Chang}, \bibinfo{person}{Kenton Lee}, {and} \bibinfo{person}{Kristina Toutanova}.} \bibinfo{year}{2019}\natexlab{}.
\newblock \bibinfo{title}{BERT: Pre-training of Deep Bidirectional Transformers for Language Understanding}.
\newblock
\newblock
\showeprint[arxiv]{1810.04805}~[cs.CL]


\bibitem[Dill(1983)]%
        {sisterhood}
\bibfield{author}{\bibinfo{person}{Bonnie~Thornton Dill}.} \bibinfo{year}{1983}\natexlab{}.
\newblock \showarticletitle{Race, Class, and Gender: Prospects for an All-Inclusive Sisterhood}.
\newblock \bibinfo{journal}{\emph{Feminist Studies}} \bibinfo{volume}{9}, \bibinfo{number}{1} (\bibinfo{year}{1983}), \bibinfo{pages}{131--150}.
\newblock
\showISSN{00463663}
\urldef\tempurl%
\url{http://www.jstor.org/stable/3177687}
\showURL{%
\tempurl}


\bibitem[Farahi(2021)]%
        {farahi2021can}
\bibfield{author}{\bibinfo{person}{Behnaz Farahi}.} \bibinfo{year}{2021}\natexlab{}.
\newblock \showarticletitle{" Can the subaltern speak?" critical making in design}.
\newblock In \bibinfo{booktitle}{\emph{ACM SIGGRAPH 2021 Art Gallery}}. \bibinfo{pages}{1--3}.
\newblock


\bibitem[Feng and Zhao(2022)]%
        {feng_hidden_2022}
\bibfield{author}{\bibinfo{person}{Violet~Du Feng} {and} \bibinfo{person}{Qing Zhao}.} \bibinfo{year}{2022}\natexlab{}.
\newblock \bibinfo{title}{Hidden Letters}.
\newblock
\newblock
\newblock
\shownote{{IMDb} {ID}: tt15503562 event-location: China}.


\bibitem[Field(2017)]%
        {facebook}
\bibfield{author}{\bibinfo{person}{Matthew Field}.} \bibinfo{year}{2017}\natexlab{}.
\newblock \showarticletitle{Facebook shuts down robots after they invent their own language}.
\newblock \bibinfo{journal}{\emph{The Telegraph}} (\bibinfo{year}{2017}).
\newblock
\urldef\tempurl%
\url{https://www.telegraph.co.uk/technology/2017/08/01/facebook-shuts-robots-invent-language/}
\showURL{%
\tempurl}


\bibitem[France(2017)]%
        {metoo}
\bibfield{author}{\bibinfo{person}{Lisa~Respers France}.} \bibinfo{year}{2017}\natexlab{}.
\newblock \bibinfo{title}{\#MeToo: Social media flooded with personal stories of assault}.
\newblock
\newblock
\urldef\tempurl%
\url{https://edition.cnn.com/2017/10/15/entertainment/me-too-twitter-alyssa-milano/index.html}
\showURL{%
\tempurl}


\bibitem[F.R.S.(1901)]%
        {pca}
\bibfield{author}{\bibinfo{person}{Karl~Pearson F.R.S.}} \bibinfo{year}{1901}\natexlab{}.
\newblock \showarticletitle{LIII. On lines and planes of closest fit to systems of points in space}.
\newblock \bibinfo{journal}{\emph{The London, Edinburgh, and Dublin Philosophical Magazine and Journal of Science}} \bibinfo{volume}{2}, \bibinfo{number}{11} (\bibinfo{year}{1901}), \bibinfo{pages}{559--572}.
\newblock
\urldef\tempurl%
\url{https://doi.org/10.1080/14786440109462720}
\showDOI{\tempurl}


\bibitem[Gao et~al\mbox{.}(2023)]%
        {gao2023symbiotic}
\bibfield{author}{\bibinfo{person}{Ze Gao}, \bibinfo{person}{Simin Yang}, \bibinfo{person}{Xingxing Yang}, {and} \bibinfo{person}{Hui Pan}.} \bibinfo{year}{2023}\natexlab{}.
\newblock \showarticletitle{Symbiotic Hands: A virtual reality interactive system that traverses reality}. In \bibinfo{booktitle}{\emph{Proceedings of EVA London 2023}}. BCS Learning \& Development, \bibinfo{pages}{163--165}.
\newblock


\bibitem[Gosse(2005)]%
        {NYradical}
\bibfield{author}{\bibinfo{person}{Van Gosse}.} \bibinfo{year}{2005}\natexlab{}.
\newblock \bibinfo{booktitle}{\emph{New York Radical Women}}.
\newblock \bibinfo{publisher}{Palgrave Macmillan US}, \bibinfo{address}{New York}, \bibinfo{pages}{123--124}.
\newblock
\showISBNx{978-1-137-04781-6}
\urldef\tempurl%
\url{https://doi.org/10.1007/978-1-137-04781-6_34}
\showDOI{\tempurl}


\bibitem[Jackson(2002)]%
        {jackson_lord_2002}
\bibfield{author}{\bibinfo{person}{Peter Jackson}.} \bibinfo{year}{2002}\natexlab{}.
\newblock \bibinfo{title}{The Lord of the Rings: The Fellowship of the Ring}.
\newblock
\newblock
\newblock
\shownote{{IMDb} {ID}: tt0120737 event-location: New Zealand, United States}.


\bibitem[Li et~al\mbox{.}(2022)]%
        {blip}
\bibfield{author}{\bibinfo{person}{Junnan Li}, \bibinfo{person}{Dongxu Li}, \bibinfo{person}{Caiming Xiong}, {and} \bibinfo{person}{Steven Hoi}.} \bibinfo{year}{2022}\natexlab{}.
\newblock \bibinfo{title}{BLIP: Bootstrapping Language-Image Pre-training for Unified Vision-Language Understanding and Generation}.
\newblock
\newblock
\showeprint[arxiv]{2201.12086}~[cs.CV]


\bibitem[Mordatch and Abbeel(2018)]%
        {mordatch2018emergence}
\bibfield{author}{\bibinfo{person}{Igor Mordatch} {and} \bibinfo{person}{Pieter Abbeel}.} \bibinfo{year}{2018}\natexlab{}.
\newblock \bibinfo{title}{Emergence of Grounded Compositional Language in Multi-Agent Populations}.
\newblock
\newblock
\showeprint[arxiv]{1703.04908}~[cs.AI]


\bibitem[M{\"u}llner(2011)]%
        {aggCluster}
\bibfield{author}{\bibinfo{person}{Daniel M{\"u}llner}.} \bibinfo{year}{2011}\natexlab{}.
\newblock \showarticletitle{Modern hierarchical, agglomerative clustering algorithms}.
\newblock \bibinfo{journal}{\emph{arXiv preprint arXiv:1109.2378}} (\bibinfo{year}{2011}).
\newblock


\bibitem[OpenAI(2023)]%
        {GPT4}
\bibfield{author}{\bibinfo{person}{OpenAI}.} \bibinfo{year}{2023}\natexlab{}.
\newblock \bibinfo{title}{GPT-4}.
\newblock
\newblock
\urldef\tempurl%
\url{https://openai.com/research/gpt-4}
\showURL{%
\tempurl}


\bibitem[Plant(1997)]%
        {plant1997zeros}
\bibfield{author}{\bibinfo{person}{S. Plant}.} \bibinfo{year}{1997}\natexlab{}.
\newblock \bibinfo{booktitle}{\emph{Zeros + Ones: Digital Women + the New Technoculture}}.
\newblock \bibinfo{publisher}{Doubleday}.
\newblock
\showISBNx{9780385482608}
\showLCCN{97010916}
\urldef\tempurl%
\url{https://books.google.co.uk/books?id=AEi0AAAAIAAJ}
\showURL{%
\tempurl}


\bibitem[Ridler and Jelonek(2017)]%
        {ridler_alice_nodate}
\bibfield{author}{\bibinfo{person}{Anna Ridler} {and} \bibinfo{person}{Daria Jelonek}.} \bibinfo{year}{2017}\natexlab{}.
\newblock \bibinfo{booktitle}{\emph{Alice \& Bob, 2017}}.
\newblock
\urldef\tempurl%
\url{http://annaridler.com/quantum-computing-art}
\showURL{%
\tempurl}


\bibitem[Sun et~al\mbox{.}(2023a)]%
        {sun2023inspire}
\bibfield{author}{\bibinfo{person}{Yuqian Sun}, \bibinfo{person}{Xingyu Li}, {and} \bibinfo{person}{Ze Gao}.} \bibinfo{year}{2023}\natexlab{a}.
\newblock \showarticletitle{Inspire creativity with ORIBA: Transform Artists' Original Characters into Chatbots through Large Language Model}.
\newblock \bibinfo{journal}{\emph{arXiv preprint arXiv:2306.09776}} (\bibinfo{year}{2023}).
\newblock


\bibitem[Sun et~al\mbox{.}(2023b)]%
        {sun2023language}
\bibfield{author}{\bibinfo{person}{Yuqian Sun}, \bibinfo{person}{Zhouyi Li}, \bibinfo{person}{Ke Fang}, \bibinfo{person}{Chang~Hee Lee}, {and} \bibinfo{person}{Ali Asadipour}.} \bibinfo{year}{2023}\natexlab{b}.
\newblock \bibinfo{title}{Language as Reality: A Co-Creative Storytelling Game Experience in 1001 Nights using Generative AI}.
\newblock
\newblock
\showeprint[arxiv]{2308.12915}~[cs.HC]


\bibitem[Sun et~al\mbox{.}(2022)]%
        {wander10}
\bibfield{author}{\bibinfo{person}{Yuqian Sun}, \bibinfo{person}{Ying Xu}, \bibinfo{person}{Chenhang Cheng}, \bibinfo{person}{Yihua Li}, \bibinfo{person}{Chang~Hee Lee}, {and} \bibinfo{person}{Ali Asadipour}.} \bibinfo{year}{2022}\natexlab{}.
\newblock \showarticletitle{Travel with Wander in the Metaverse: An AI chatbot to Visit the Future Earth}. In \bibinfo{booktitle}{\emph{2022 IEEE 24th International Workshop on Multimedia Signal Processing (MMSP)}}. \bibinfo{pages}{1--6}.
\newblock
\urldef\tempurl%
\url{https://doi.org/10.1109/MMSP55362.2022.9950031}
\showDOI{\tempurl}


\bibitem[Sun et~al\mbox{.}(2023c)]%
        {wander20}
\bibfield{author}{\bibinfo{person}{Yuqian Sun}, \bibinfo{person}{Ying Xu}, \bibinfo{person}{Chenhang Cheng}, \bibinfo{person}{Yihua Li}, \bibinfo{person}{Chang~Hee Lee}, {and} \bibinfo{person}{Ali Asadipour}.} \bibinfo{year}{2023}\natexlab{c}.
\newblock \showarticletitle{Explore the Future Earth with Wander 2.0: AI Chatbot Driven By Knowledge-Base Story Generation and Text-to-Image Model}. In \bibinfo{booktitle}{\emph{Extended Abstracts of the 2023 CHI Conference on Human Factors in Computing Systems}} (Hamburg, Germany) \emph{(\bibinfo{series}{CHI EA '23})}. \bibinfo{publisher}{Association for Computing Machinery}, \bibinfo{address}{New York, NY, USA}, Article \bibinfo{articleno}{450}, \bibinfo{numpages}{5}~pages.
\newblock
\showISBNx{9781450394222}
\urldef\tempurl%
\url{https://doi.org/10.1145/3544549.3583931}
\showDOI{\tempurl}


\bibitem[Wajcman(1991)]%
        {wajcman1991feminism}
\bibfield{author}{\bibinfo{person}{J. Wajcman}.} \bibinfo{year}{1991}\natexlab{}.
\newblock \bibinfo{booktitle}{\emph{Feminism Confronts Technology}}.
\newblock \bibinfo{publisher}{Pennsylvania State University Press}.
\newblock
\showISBNx{9780271008028}
\showLCCN{91018539}
\urldef\tempurl%
\url{https://books.google.co.uk/books?id=BtaiFSv09jMC}
\showURL{%
\tempurl}


\bibitem[Wang(2011)]%
        {wang_snow_2011}
\bibfield{author}{\bibinfo{person}{Wayne Wang}.} \bibinfo{year}{2011}\natexlab{}.
\newblock \bibinfo{title}{Snow Flower and the Secret Fan}.
\newblock
\newblock
\newblock
\shownote{{IMDb} {ID}: tt1541995 event-location: China, United States}.


\bibitem[Yang et~al\mbox{.}(2023)]%
        {yang2023tangible}
\bibfield{author}{\bibinfo{person}{Simin Yang}, \bibinfo{person}{Ze Gao}, \bibinfo{person}{Reza Hadi~Mogavi}, \bibinfo{person}{Pan Hui}, {and} \bibinfo{person}{Tristan Braud}.} \bibinfo{year}{2023}\natexlab{}.
\newblock \showarticletitle{Tangible Web: An Interactive Immersion Virtual Reality Creativity System that Travels Across Reality}. In \bibinfo{booktitle}{\emph{Proceedings of the ACM Web Conference 2023}}. \bibinfo{pages}{3915--3922}.
\newblock


\bibitem[Zhang et~al\mbox{.}(2021)]%
        {zhang2021cangjie}
\bibfield{author}{\bibinfo{person}{Weidi Zhang}, \bibinfo{person}{Donghao Ren}, {and} \bibinfo{person}{George Legrady}.} \bibinfo{year}{2021}\natexlab{}.
\newblock \showarticletitle{Cangjie's poetry: an interactive art experience of a semantic human-machine reality}.
\newblock \bibinfo{journal}{\emph{Proceedings of the ACM on Computer Graphics and Interactive Techniques}} \bibinfo{volume}{4}, \bibinfo{number}{2} (\bibinfo{year}{2021}), \bibinfo{pages}{1--9}.
\newblock


\bibitem[Zhao and Zhang(1995)]%
        {ZhaoNushu}
\bibfield{author}{\bibinfo{person}{Liming Zhao} {and} \bibinfo{person}{Gongjin Zhang}.} \bibinfo{year}{1995}\natexlab{}.
\newblock \bibinfo{booktitle}{\emph{N$\ddot{\mathrm{u}}$ Shu Yu N$\ddot{\mathrm{u}}$ Shu Wen Hua}}.
\newblock \bibinfo{publisher}{Xin hua chu ban she}.
\newblock


\end{thebibliography}

\end{document}